\documentclass[letterpaper, 10 pt, conference]{ieeeconf}  

\IEEEoverridecommandlockouts                              

\usepackage{booktabs}       
\usepackage{graphicx}
\usepackage{amsmath}   
\usepackage{amssymb}   
\usepackage{multirow}
\usepackage{subcaption}
\usepackage{makecell}
\usepackage{hyperref}
\usepackage{xcolor}

\title{\LARGE \bf
Look, Focus, Act: Efficient and Robust Robot Learning via\\ Human Gaze and Foveated Vision Transformers
}

\author{Ian Chuang$^{1}$ Jinyu Zou$^{2}$ Andrew Lee$^{3}$ Dechen Gao$^{3}$ Iman Soltani$^\dagger$$^{3}$
\thanks{
$^{1}$University of California, Berkeley 
$^{2}$Tongji University}
\thanks{$^{3}$University of California, Davis}
\thanks{$^{\dagger}$Corresponding email: \href{mailto:isoltani@ucdavis.edu}{isoltani@ucdavis.edu}}
}

\begin{document}

\maketitle
\thispagestyle{empty}
\pagestyle{empty}

\begin{abstract}

Human vision is a highly active process driven by gaze, which directs attention to task-relevant regions through foveation, dramatically reducing visual processing. In contrast, robot learning systems typically rely on passive, uniform processing of raw camera images. In this work, we explore how incorporating human-like active gaze into robotic policies can enhance efficiency and robustness. We develop GIAVA (Gaze Integrated Active-Vision ALOHA), a robot vision system that emulates human head and neck movement, and gaze adjustment for foveated processing. Extending the AV-ALOHA robot platform, we introduce a framework for simultaneously collecting eye-tracking, perspective control, and robot manipulation demonstration data from a human operator. We also open-source a simulation benchmark and dataset for training robot policies that incorporate human gaze. Inspired by recent work in foveated image segmentation and given the widespread use of Vision Transformers (ViTs) in robot learning, we integrate gaze information into ViTs using a foveated patch tokenization scheme. Compared to uniform patch tokenization, this significantly reduces the number of tokens, and thus computation. For this purpose, we explore two approaches to gaze estimation: The first is a two-stage model that predicts gaze independently to guide foveation and subsequently action. The second integrates gaze into the action space, allowing the policy to jointly estimate gaze and actions end-to-end. Our results show that our method for foveated robot vision drastically reduces computational overhead, and enhances robustness to background distractors. Notably, on certain high-precision tasks, foveated vision also improves performance, as reflected in higher success rates. Together, these findings suggest that human-inspired foveated visual processing offers untapped potential and should be further considered as a useful inductive bias in robotic vision systems. \href{https://soltanilara.github.io/giava/}{\textcolor{blue}{https://soltanilara.github.io/giava/}}

\end{abstract}

\section{Introduction}

\begin{figure}[t]
    \centering
    \includegraphics[width=\linewidth]{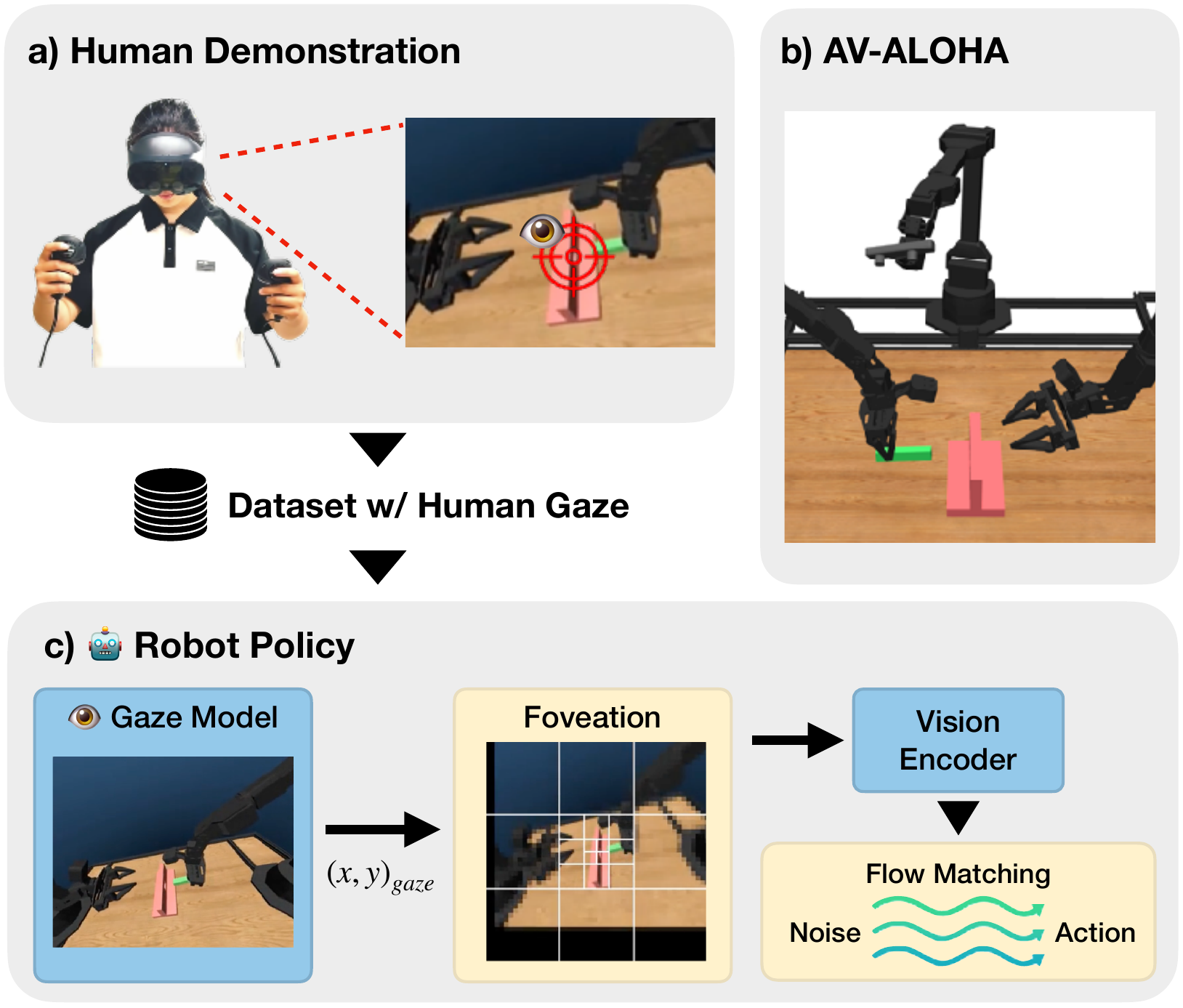}
    \caption{\textbf{GIAVA System Overview:} a) Bimanual demonstrations with eye-tracking are recorded using a VR headset to b) control the AV-ALOHA robot system. After data collection, a robot policy is trained to imitate human gaze, head/neck movement and manipulation actions. c) The policy first predicts gaze, foveates the image around the gaze, and generates actions via a flow matching policy.}
    \label{fig:av-aloha}
    \vspace{-15pt}
\end{figure}

Imitation Learning (IL) has emerged as a powerful approach to enabling dexterous robot behaviors in complex systems, such as bimanual manipulation \cite{zhao2023learning, zhao2024aloha, lee2024interact, gao2025vita} and humanoid control \cite{ze2024generalizable, fmp-multisupport, he2024hover}. These methods typically process camera images and robot proprioception to directly produce robot actions end-to-end \cite{chi2023diffusion}. However, despite their goal of mimicking human demonstrations, most methods process visual input in ways that diverge markedly from how humans perceive and use visual information.

Unlike typical robotic vision systems that process the entire visual input uniformly, the human visual system is optimized for efficiency through foveation~\cite{schira2009foveal, wang2001embedded}. High spatial resolution vision is concentrated in the fovea, which occupies a disproportionately large portion of the primary visual cortex. Replicating this high resolution uniformly across the whole visual field would require an approximately 1000x increase in the size of the primary visual cortex~\cite{10.1371/journal.pcbi.1005743}. This selective focus not only reduces metabolic cost, but also enables humans to allocate cognitive resources efficiently. Understanding and modeling human gaze behavior, particularly how the fovea is directed toward regions of interest, can offer valuable insights for robot learning. We hypothesize that leveraging gaze and foveation to guide a robot’s visual attention enables more efficient perception and possibly more effective action in complex tasks and environments. 

To incorporate gaze into IL, one can leverage increasingly popular Virtual Reality (VR) headsets for collecting demonstrations \cite{chuang2024active, cheng2024open}. Modern VR headsets often feature built-in eye-tracking capabilities, allowing simultaneous recording of gaze data and robot demonstrations. Capturing gaze data alongside motor actions offers valuable supervision, providing insights into where to focus attention. By learning not only from a demonstrator’s actions but also from their visual attention, we can take steps toward developing robotic vision systems that more closely emulate human visual efficiency and potentially human-level dexterity. 

This also motivates a shift in how we design visual processing systems for robot learning. Rather than encoding entire images uniformly, we show that successful task execution can simply rely on attending to a few key regions of interest. This is particularly relevant for Vision Transformers (ViTs)~\cite{dosovitskiy2020image}, which are widely used in robot learning due to their strong capabilities in visual representation learning \cite{oquab2023dinov2, cheng2024open, lin2024data, xia2024cage}. ViTs are costly since they compute pairwise interactions among all spatial tokens using self attention, which yields $O(N^2)$ complexity in the number of tokens. At the same time, the token-based and spatially agnostic design that incurs this cost makes ViTs well suited to foveated vision, because token density and attention can be concentrated near the fixation point and sparsified in the periphery. We hypothesize that by conditioning tokenization and attention on gaze, which presumably points to the most task-relevant part of the image, we can preserve, or even improve, performance while reducing computational load.

Despite the biological relevance of gaze and foveation and the increasing accessibility of eye-tracking, their integration into robot learning remains limited. To address this gap, we propose a system that combines recent advances in IL and foveated visual processing as illustrated in Fig.~\ref{fig:av-aloha}. 

First, we introduce GIAVA, a robot platform that enables efficient and accessible collection of human demonstrations and eye-tracking data using a VR headset. Building on AV-ALOHA \cite{chuang2024active}, which enabled robots to learn active vision (AV), i.e., camera viewpoint control, from human demonstrations, we extend the framework to also learn human gaze behavior. We also release GIAVA's simulation replica along with open-source datasets containing synchronized human demonstrations and eye-tracking data. This benchmark aims to support the community in exploring how best to leverage both gaze and active vision for IL.

Second, we introduce an IL policy that integrates gaze estimation and foveated image processing. We explore two approaches to gaze estimation: a) a hierarchical, modular strategy where the robot first identifies where to look before deciding how to act, and b) an end-to-end approach that jointly predicts gaze and robot actions. In both approaches, we integrate the generated gaze with a foveated ViT method from \cite{schmidt2025segment}, originally developed for image segmentation, for use in IL. This method uses ``foveated tokenization'' which allocates high-resolution patches near the gaze point and coarser patches towards the periphery, mirroring the human retina’s photoreceptor density reduction from central towards peripheral vision \cite{mehri2017non}. We find that incorporating gaze and foveation improves robustness to distractors, while significantly reducing the computational cost.

Our contributions can be summarized as the following:

1) \textbf{Biologically-Inspired Foveated Vision System:} We demonstrate the potential of foveated ViTs for robot learning. This approach mimics human vision by concentrating patches at a predicted gaze point, which reduces the number of visual tokens and the associated ViT computation by 94\% while preserving or potentially improving performance. 

2) \textbf{Gaze-Enhanced Policy Learning Framework:} We propose and evaluate two distinct methods for integrating gaze with IL policies. The first is a hierarchical, two-stage approach that first estimates gaze and then uses it as inputs to the policy. The second is a novel end-to-end method that concurrently generates gaze and action via a shared policy network, treating gaze as part of the robot's action space.

3) \textbf{Public Benchmark and Dataset with Extensive Experiments:} We demonstrate through extensive experiments that our foveated approach enhances robustness and can also improve performance for certain tasks, all while reducing computational complexity by 94\%, speeding up training by 7x and inference by 3x. To facilitate further research, we open-source the GIAVA platform and datasets.

\section{Related Work}

\subsection{Biologically-Inspired Visual Processing}

The study of human vision has been an active area of research for decades across neuroscience, psychology, and cognitive science~\cite{marr2010vision, biederman1987recognition, desimone1995neural, dicarlo2012does, yamins2014performance}. With the emergence of deep learning and embodied AI, there has been growing interest in understanding and incorporating biologically-inspired principles such as foveation or selective attention into artificial visual processing systems~\cite{wijntjes2018context, deza2020emergent}. A number of works leverage foveated representations to reduce redundant computation and focus model capacity on salient image regions, specifically incorporating foveation into Convolutional Neural Networks (CNNs)~\cite{lukanov2021biologically, jonnalagadda2021foveater}. More recently, ViTs have extended these ideas into transformer-based architectures. For example, Peripheral ViT~\cite{min2022peripheral} introduces biologically-inspired positional encodings that mimic the human retina's spatially varying resolution. Segment This Thing~\cite{schmidt2025segment} uses a variable resolution patch tokenization pattern centered around a point prompt for use in image segmentation, reducing computational cost while preserving accuracy. Inspired by this design, we incorporate a gaze-guided foveated tokenization method into robot learning, enabling more efficient and human-like visual processing. 

\subsection{Gaze for Robotics}

Although still an emerging area in robotics, a few studies have investigated how human gaze can support robotic perception, control, and human-robot interaction~\cite{carreto2018eye, zhang2023human, belardinelli2024gaze}. Recent robot learning works have also leveraged gaze-inspired methods in the context of reinforcement learning (RL). ViSaRL~\cite{liang2024visarl} uses manually annotated saliency maps to pretrain visual representations, which is then used to improve downstream RL performance. EyeRobot~\cite{kerr2025eye}, on the other hand, learns gaze behavior through RL by optimizing a task-driven reward that encourages eye movements which improve the performance of a co-trained manipulation policy. Although neither approach uses actual human gaze, both works highlight the value of gaze-like signals in shaping perception for effective robot learning. 

IL provides a more direct way to investigate the utility of foveated visual processing in robotics through imitating human gaze. One initial effort in this line of work used Mixture Density Models to predict gaze points from human data and crop the corresponding regions of interest, which were then fed into a policy \cite{heecheol2020}. Subsequent works used gaze to switch from low-resolution full images to high-resolution crops for actions that require more precision~\cite{kim2021gaze}, and from local end-effector actions to global reaching actions depending on gaze direction~\cite{kim2024multi}. 

Unlike existing approaches, our method a) explores a more sophisticated foveated image processing method for robot learning leveraging patch processing architecture of ViTs, and b) adopts a more human-like hardware setup in which gaze adjustment is integrated with a 7-DoF robotic arm representing head and neck movement for perspective control. We also provide learning and architectural recipes, releasing an open-source simulation and hardware stack to facilitate further investigation by the community.

\section{Method}

\subsection{Data Collection with Eye Tracking}

\begin{figure}[t]
    \centering
    \includegraphics[width=\linewidth]{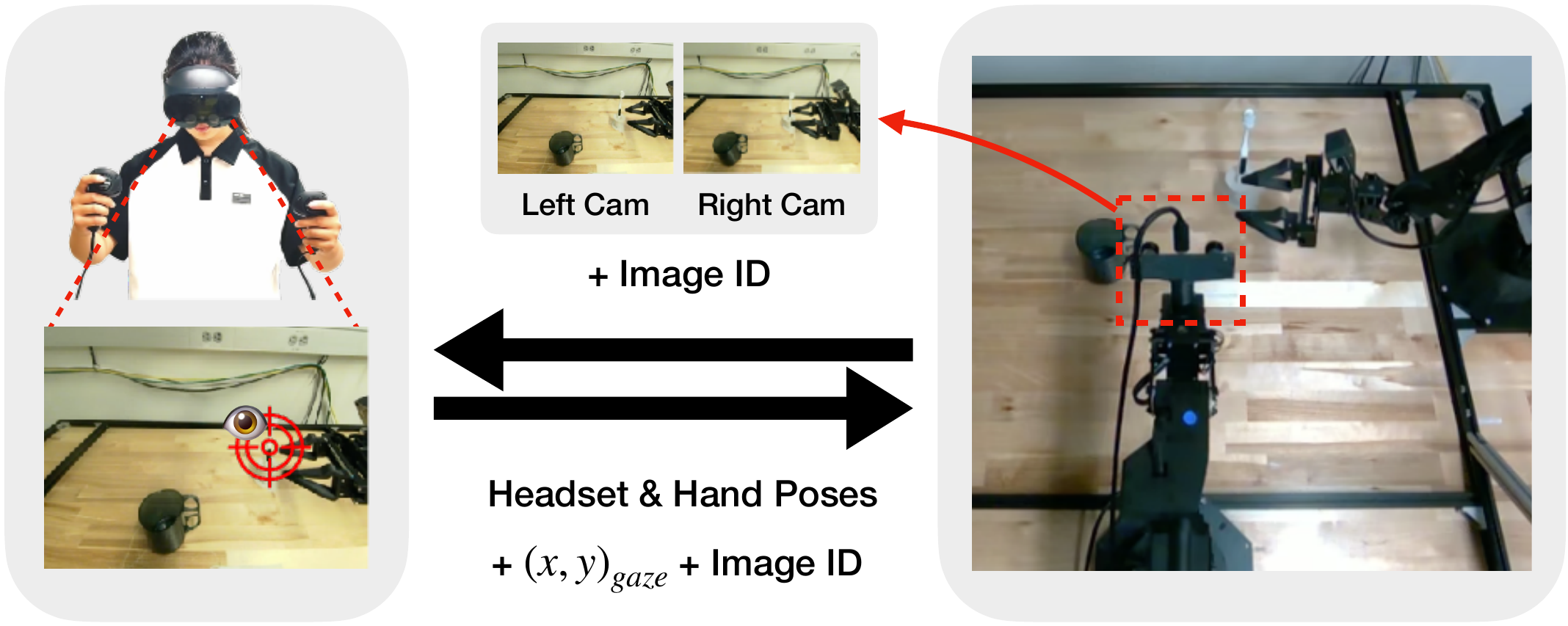}
    \caption{\textbf{GIAVA Data Collection:} The stereo camera images along with an image ID are transmited to the VR headset. The VR headset sends back head and hand controller poses to control the manipulation and active vision robotic arms, along with gaze coordinates and the corresponding image ID, synchronizing gaze data with the images.}
    \label{fig:headset_communication}
    \vspace{-15pt}
\end{figure}

To collect robot demonstration data with eye tracking, we extend AV-ALOHA \cite{chuang2024active} to the GIAVA setup. AV-ALOHA builds on the original ALOHA platform \cite{zhao2023learning} (featuring two robotic manipulation arms), by adding a third 7-DOF AV arm equipped with a stereo camera to dynamically adjust its viewpoint during task execution. The AV robot is operated by the real-time pose of a VR-headset worn by a user. Manipulation arms are controlled by VR controllers, enabling simultaneous control of all three robot arms via head and hand movements. The user receives real-time visual feedback through a video stream from the robot's stereo camera to the headset display. To collect eye tracking data, we replace the Meta Quest 3 headset used in the original setup with the Meta Quest Pro, which includes built-in eye tracking sensors. We also upgrade the stereo camera to increase the vertical field of view from 52$^{\circ}$ to 105$^{\circ}$, improving visibility for gaze tracking. These changes allow for easy recording of gaze data along with viewpoint control and manipulation demonstrations.

The VR headset streams head and hand pose data to the robot, which are then converted to joint commands through inverse kinematics. The robot streams images from its stereo camera to the headset, which are displayed to the user's left and right eye to provide depth perception. The headset also transmits gaze data, specifically, the image coordinates of the user's left and right gaze points on the corresponding camera image, which are recorded by the robot. During data collection, we synchronously record the robot's camera images, joint states, actions, and human gaze coordinates.

To address the potential latency and resulting misalignment between eye-tracking and images, we annotate each image frame sent from the robot to the VR headset with a unique ID. When the VR headset streams head and hand pose data to the robot, it also sends eye-tracking data tagged with the corresponding image ID, ensuring proper synchronization. For images that are not labeled in time with the corresponding gaze, we interpolate between known eye-tracking labels to approximate the gaze data. Fig.~\ref{fig:headset_communication} illustrates the details of the information exchange process.

\subsection{Flow Matching Policy}

\begin{figure*}[t]
    \centering
    \includegraphics[width=1.0\textwidth]{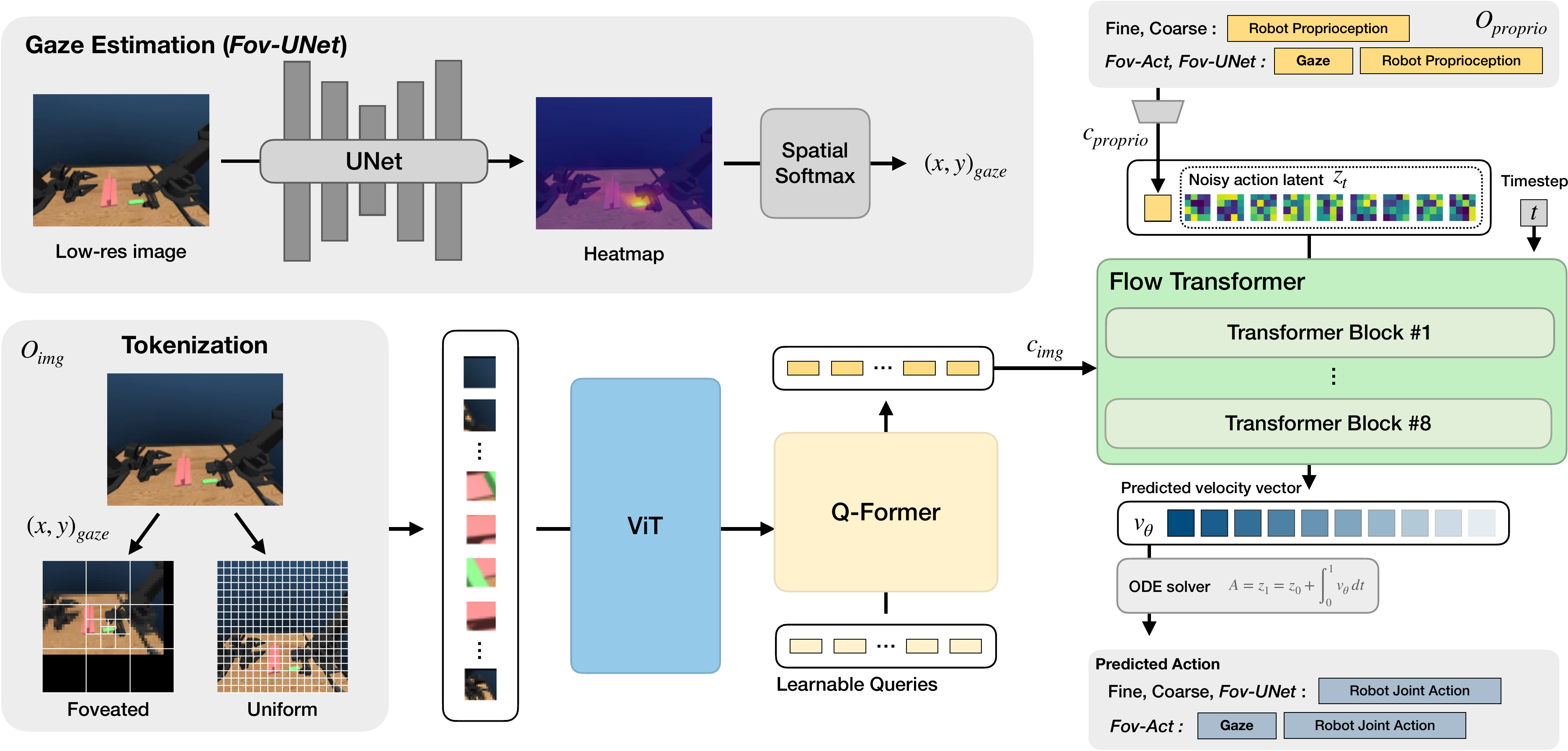}
    \caption{ \textbf{Gaze Estimation:} \textit{Fov-UNet} uses UNet and spatial softmax to predict gaze; \textit{Fov-Act} predicts future gaze and action together via policy. \textbf{Tokenization:} \textit{Fov-UNet} and \textit{Fov-Act} use foveated tokenization around predicted gaze; \textit{Fine} and \textit{Coarse} use uniform tokenization. \textbf{Policy Architecture:} Image observations $O_{\text{img}}$ are tokenized, processed by ViT, and compressed with Q-Former module into tokens $c_{\text{img}}$, which conditions Flow Transformer (FT) via cross-attention. Proprioception is encoded by MLP into tokens $c_{\text{proprio}}$ and added to FT input sequence. Timestep $t$ is embedded and conditions FT via AdaLN. FT predicts velocity $v_{\theta}$ from noisy action latent $z_t$, $c_{\text{img}}$, $c_{\text{proprio}}$, and $t$. Actions are generated via Euler integration.}
    \label{fig:architecture}
    \vspace{-15pt}
\end{figure*}

The policy, denoted as $\pi(A|O)$, maps an observation $O$ to an action chunk $A$ of length $K$ (we use $K=16$ in experiments). We apply conditional flow matching (CFM) \cite{tong2023improving, zhang2024affordance} to learn the policy from expert demonstrations. Flow matching learns a time-dependent vector field $v_{\theta}(z_t, t, O)$ that models the flow from a simple prior distribution, typically a normal distribution $p_0 = \mathcal{N}(0, I)$, to the target conditional distribution of expert actions, $p_1(A|O)$. The path between a sample $z_0 \sim p_0$ and a corresponding action $A \sim p_1(A|O)$ is defined by the velocity field $u_t(z|A, z_0)$. The model parameters $\theta$ are optimized by minimizing the mean squared error between the predicted velocity $v_{\theta}$ and the ground-truth velocity $A - z_0$. The loss is an expectation over the time step $t \in [0, 1]$, the observation $O$, the target action $A$, and the latent variable $z_t$ sampled along the path: $
\mathcal{L}_{\text{CFM}}(\theta) 
= \mathbb{E}_{t, (A,O), z_0 \sim p_0}\Big[
  \big\| v_{\theta}((1-t)z_0 + tA, t, O) - u_t(z|A, z_0)\big\|^2 
\Big] $. To generate an action chunk during inference, we first draw a random noise vector $z_0 \sim \mathcal{N}(0, I)$. We then solve the learned ordinary differential equation (ODE)~\cite{lipman2022flow}, $\frac{dz_t}{dt} = v_{\theta}(z_t, t, O)$, by numerically integrating from $t=0$ to $t=1$. The resulting state $z_1$ is the generated chunk $A$. We use a simple Euler ODE solver with 8 discretization steps to approximate for this integration.

\subsection{Policy Architecture}

The model architecture, outlined in Fig~\ref{fig:architecture}, is designed to process inputs using a ViT based ~\cite{dosovitskiy2020image} observation encoder, and generate actions using a transformer based action decoder, which parameterizes $v_{\theta}(z_t, t, O)$.


\textbf{Observation Encoder}: The raw observation $O$ consists of an image from the robot's left eye (i.e., the left camera of the stereo camera mounted on GIAVA’s active vision arm), $O_{\text{img}}$, and its proprioceptive state (joint angles), $O_{\text{proprio}}$. The image is passed through the ViT, which outputs a sequence of feature tokens~\cite{dosovitskiy2020image}. The ViT tokens are then processed by a Q-Former module \cite{xia2024cage, li2023blip}. This module uses a small set of 16 learnable queries to distill the extensive visual information into a compact set of conditioning tokens, $c_{\text{img}}$, via cross-attention. The robot's proprioceptive input is encoded using a multi-layer perceptron (MLP) to project to the token dimension, yielding $c_{\text{proprio}}$. 

\textbf{Action Decoder}: The core of our policy is a transformer that learns the velocity field $v_{\theta}$ for flow matching. Its input is a sequence of tokens representing the action latent $z_t$. In addition, $c_{\text{proprio}}$ is concatenated to the transformer's input sequence for conditioning. The transformer is structured as multiple AdaLN-Zero blocks \cite{peebles2023scalable, polyak2024movie}. Specifically, to condition the time step $t$, it is embedded then injected through Adaptive Layer Norm (AdaLN) for each block. In addition, each AdaLN-Zero block includes a cross-attention layer where the action-proprioception sequence attends to the image features $c_{\text{img}}$, allowing the model to integrate visual information at every stage of processing. The transformer predicts a velocity vector $v_{\theta}(z_t, t, O)$, which is used for both the training objective and the inference-time ODE solver.

\subsection{Gaze Estimation}

While human gaze is available during training, it is not accessible at test time. The policy must therefore learn to estimate gaze in addition to robot actions. We propose two approaches: a two-stage method and an end-to-end method.

In the two-stage method ~\cite{kim2021gaze, kim2024multi}, we first estimate gaze, then use the estimated gaze to guide foveation and condition the action policy. A downscaled camera image is processed by a UNet with a ResNet18 backbone pretrained on ImageNet~\cite{deng2009imagenet}, producing a heatmap. A spatial softmax then extracts a gaze keypoint from the heatmap, which is supervised during training with mean squared error against the ground-truth human gaze. The gaze model is trained separately for 30,000 steps (batch size 64, learning rate $10^{-4}$) and frozen during policy training. The predicted gaze both drives the foveation process and is also appended to the robot’s proprioceptive input for policy conditioning.

In our end-to-end method, we instead treat gaze as part of whole-body control, predicting a trajectory of future gaze jointly with actions. This requires extending the policy’s action space to include gaze. The policy's gaze estimation is then used for foveation in the next inference step and is also concatenated to the robot's proprioception. In this case, an initial gaze is needed to start the sequence, which we set to the center of the image. This unifies gaze and action estimation under the flow matching framework, producing synchronized gaze-action trajectories.

We note that there are inherent trade-offs between the two methods. The two-stage approach requires an additional UNet model, which increases both training and inference time. In contrast, while the end-to-end gaze estimation method is more efficient, it directly predicts gaze keypoints, lacking the inductive spatial bias of the UNet-based model that estimates a 2D heatmap. Furthermore, due to its prediction of a future gaze based on a current image frame, it may be more prone to estimation errors, especially under dynamic settings with greater variability across consecutive frames. 

\subsection{Foveated Tokenization}

We foveate the input observation image at the estimated gaze to focus the policy on relevant regions in the image and reduce visual processing overhead. With ViTs becoming increasingly common in robot learning \cite{lin2024data, kerr2025eye, xia2024cage}, we adopt the foveated patch tokenization method introduced in \cite{schmidt2025segment} for image segmentation and adapt it for robot learning.

Unlike standard ViT tokenization, which uses a uniform grid of equally sized patches, this foveated approach mimics human vision by placing small, densely packed patches at the center, corresponding to the gaze point, and arranging larger, sparser patches in concentric rings toward the periphery. To incorporate gaze, we shift the image so that the estimated gaze aligns with the center of the foveation pattern. This ensures that the region around the gaze is represented with higher spatial resolution. As this shift moves parts of the image beyond the original boundaries, we pad the image with zeros. All patches are then downscaled to match the size of the central patches, which can then be passed to a standard ViT. For our implementation, we use a custom \textit{Foveated} pattern that uses only 20 patches. 

\begin{figure}[t]
    \centering
    \includegraphics[width=\linewidth]{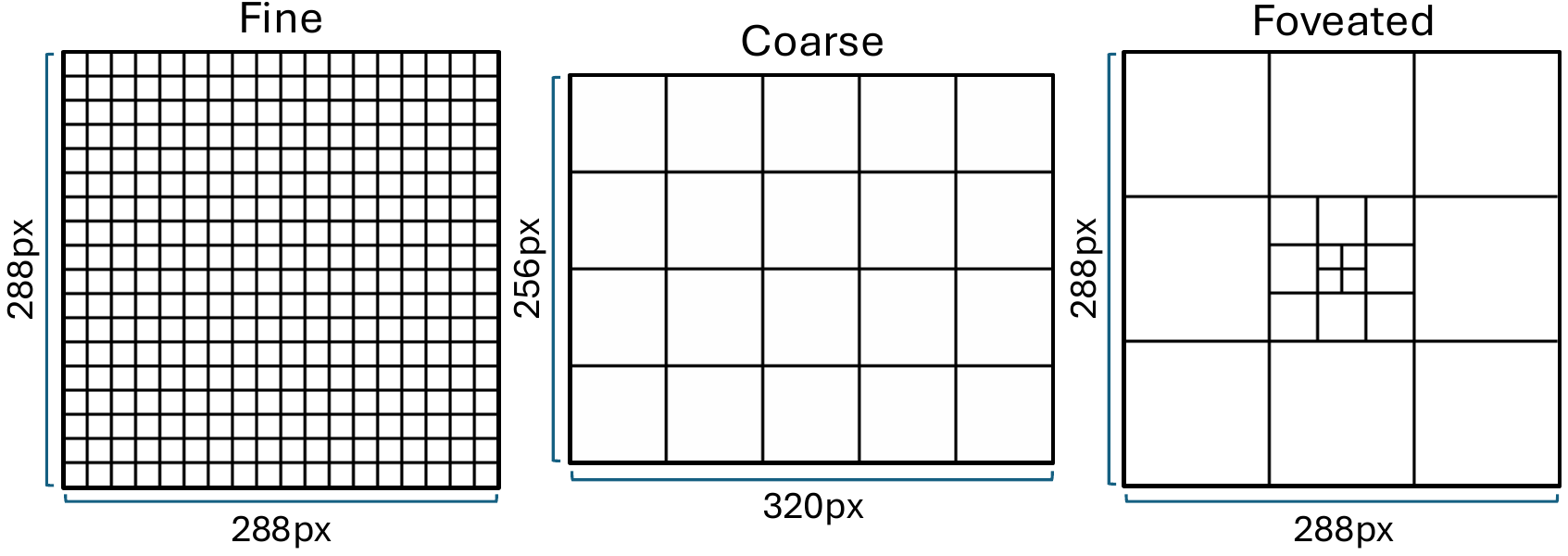}
    \caption{Visualization of the patch tokenization methods.}
    \label{fig:vit_patterns}
    \vspace{-15pt}
\end{figure}

To compare against the \textit{Foveated} tokenization pattern, we evaluate two uniform patchification strategies. The first, referred to as the \textit{Fine} pattern, uses a uniform 18×18 grid of 16×16 pixel patches, matching the size of the center patches in the foveated pattern and covering the same overall image area. This results in 324 tokens, which is 16.2 times more than the foveated pattern. The second, called the \textit{Coarse} pattern, uses the same total number of tokens as the foveated pattern (20 patches), but uniformly arranged in a 4×5 grid. Consequently, each patch covers a much larger area of 64×64 pixels. A visualization of these patterns are shown in Fig.~\ref{fig:vit_patterns}.

\begin{figure}[t]
    \centering
    \includegraphics[width=\linewidth]{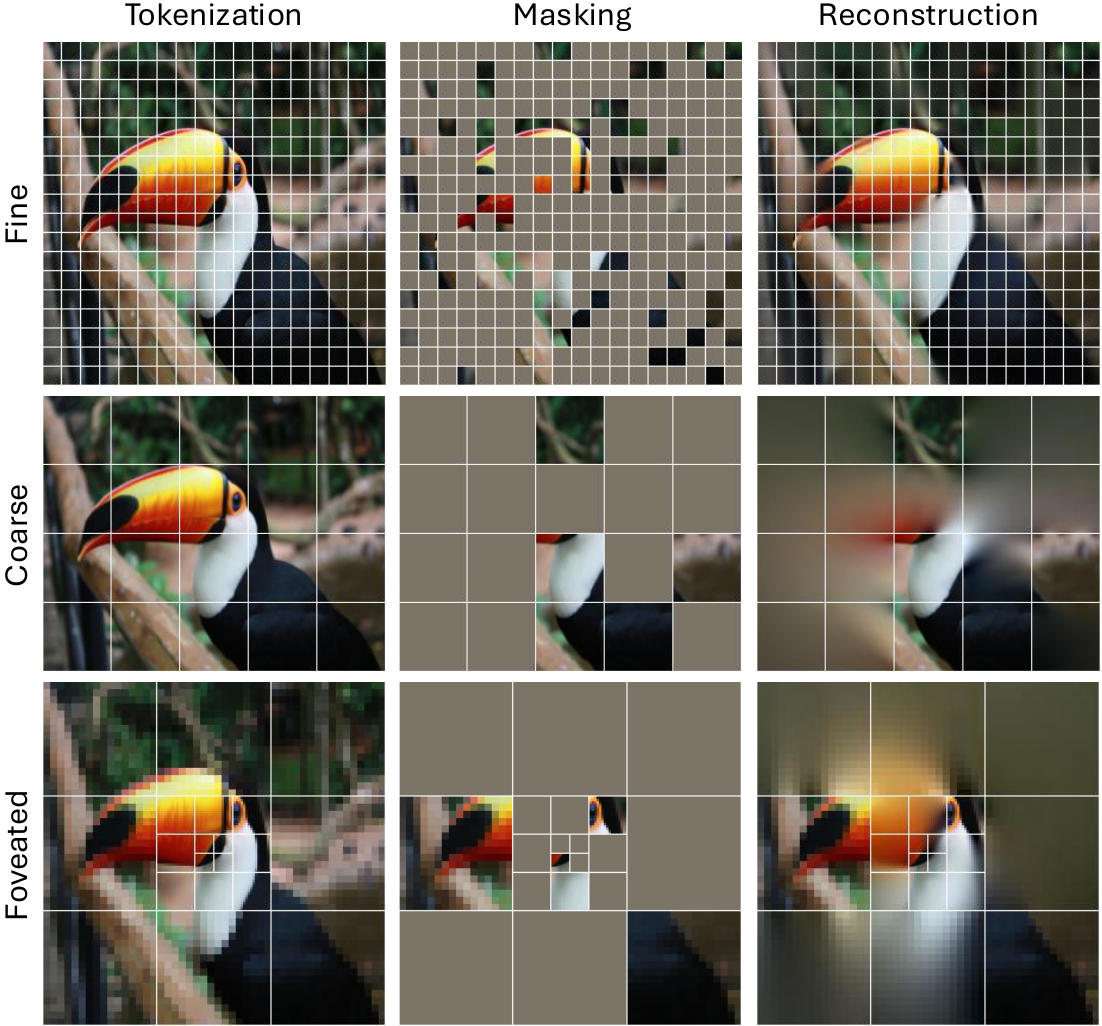}
    \caption{MAE reconstructions after pretraining with different patch tokenization patterns. The input image is tokenized (left); a subset of tokens is passed to the encoder (center); the full image is reconstructed by the decoder (right).}
    \label{fig:mae_viz}
    \vspace{-15pt}
\end{figure}

One downside of using a foveated tokenization scheme is that open-source pretrained ViT weights, commonly used in robot learning for their significant performance benefits \cite{lin2024data, xia2024cage}, cannot be directly applied as pretraining is done on fixed, uniform tokenization patterns. To address this, we pretrain our own ViT-B models from scratch using the Masked Autoencoder (MAE) objective \cite{he2022masked} for the \textit{Foveated}, \textit{Fine}, and \textit{Coarse} tokenization patterns. Due to limited computational resources, instead of training on the full ImageNet-1K dataset \cite{deng2009imagenet}, we train on a smaller subset of 60,000 images for 1,000 epochs, following the standard MAE pretraining procedure. We acknowledge that performance for the uniform tokenization patterns could improve by using popular pretrained weights such as DINOv2 \cite{oquab2023dinov2}, but to ensure a fair comparison, we use the same procedure across all patterns. An example visualization of MAE pretraining results for all three patterns is shown in Fig.~\ref{fig:mae_viz}.

\section{Experiments}

\begin{figure*}[t]
    \centering
    \includegraphics[width=1.0\textwidth]{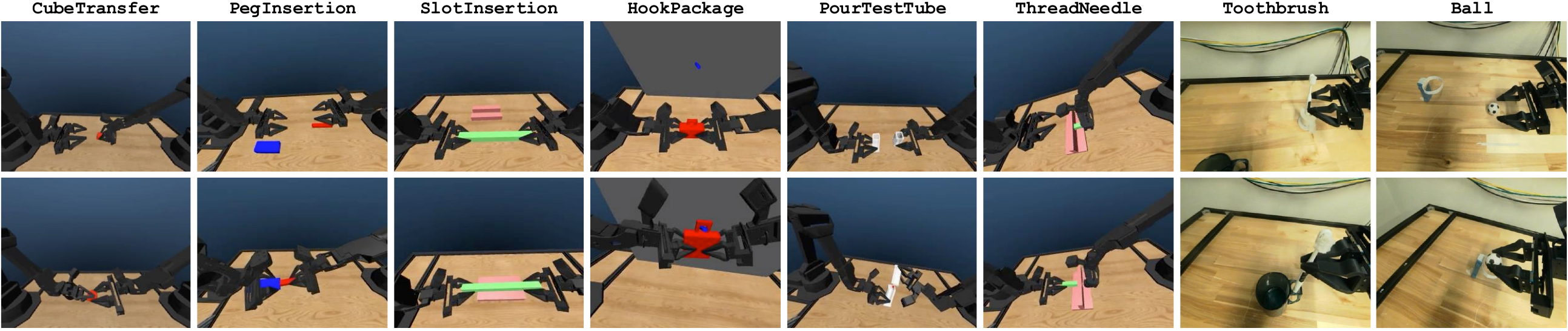}
    \caption{Illustration of simulation and real tasks.}
    \label{fig:tasks}
    \vspace{-5pt}
\end{figure*}

\begin{figure*}[htbp]
    \centering
    \includegraphics[width=1.0\textwidth]{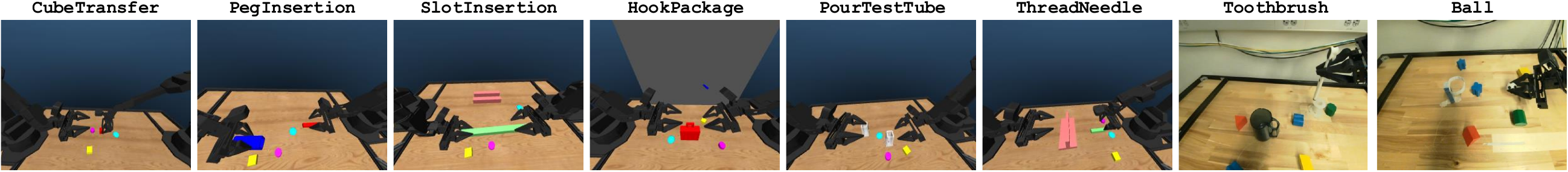}
    \caption{We evaluate all tasks with randomly placed distractor objects.}
    \label{fig:distractors}
    \vspace{-15pt}
\end{figure*}

We evaluate on six simulation tasks from the AV-ALOHA benchmark~\cite{chuang2024active} and two real-robot tasks shown in Fig.~\ref{fig:tasks}. For each simulation task, we collected 100 demonstrations with eye-tracking. For real-robot experiments, we test on two tasks: \texttt{Ball}, where the arm picks up and drops a ball into a toy hoop (60 episodes), and \texttt{Toothbrush}, where the arm inserts a toothbrush into a small hole (78 episodes). For each task, all demonstrations are collected by a single human operator. We also randomize the initial object placements during both data collection and evaluation to increase diversity in the demonstrations and test conditions.

We also introduce distractors for each task to evaluate robustness in unseen cluttered environments. For distractors, we randomly place small objects of varying colors and shapes near the primary objects to be manipulated. Examples of these distractors for each task are shown in Fig.~\ref{fig:distractors}.

In our experiments, we evaluate four policies that differ in their patch tokenization strategies and, for foveated variants, in their gaze estimation methods. The \textit{Fine} policy uses the \textit{Fine} tokenization pattern, while the \textit{Coarse} policy uses the \textit{Coarse} pattern. The \textit{Fov-Act} policy adopts the \textit{Foveated} pattern and uses the end-to-end gaze prediction method. In contrast, \textit{Fov-UNet}, which follows the \textit{Foveated} pattern, uses the UNet-generated gaze described earlier. 

We evaluate each method using both randomly initialized ViT weights and our MAE-pretrained ViT weights to assess the impact of pretraining. Additionally, we evaluate each method in two settings, without distractors (\textit{Standard}) and with distractors present in the scene (\textit{Distractors}). For real experiments, we evaluate \textit{Fine} and \textit{Fov-UNet} with MAE pretraining only. 

Each policy is trained and evaluated on a single task, with evaluations performed at 8.33 FPS for simulation and 11 FPS for the real robot. The policy predicts action chunks of size 16. We incorporate a temporal ensemble technique from \cite{zhao2023learning} to produce smoother motions and improve responsiveness at inference time. Policies are trained for 30,000 steps with a batch size of 64. The learning rate is set to $10^{-4}$; however, if MAE-pretrained ViT weights are used, the ViT learning rate is reduced to $10^{-5}$ as recommended in \cite{chi2023diffusion}. A cosine learning rate scheduler is used, along with exponential moving average (EMA) updates with a decay rate of 0.99. For real-robot tests, during training we further applied data augmentations, such as Gaussian blur and color jitter, to compensate for limited pretraining.  

\section{Results}

\subsection{Efficiency}

\begin{table}[t]
\centering
\begin{tabular}{l|cc|cc}
\toprule
& \multicolumn{2}{c|}{\textbf{Training}} & \multicolumn{2}{c}{\textbf{Inference}} \\
\cmidrule(lr){2-3} \cmidrule(lr){4-5}
\textbf{Policy} & \textbf{\makecell{Latency \\ (ms/step)}} & \textbf{\makecell{Memory \\ (MiB)}} & \textbf{\makecell{Latency \\ (ms/chunk)}} & \textbf{\makecell{Memory \\ (MiB)}} \\
\midrule
\textbf{Fine}     & 833.2 & 20949 & 334.7 & 2281 \\
\textbf{Coarse}   & 109.6 & 4083 & 89.1 & \textbf{1327} \\
\textbf{Fov-Act}  & \textbf{108.2} & \textbf{3937} & \textbf{87.9} & 1435 \\
\textbf{Fov-UNet} & 123.8 & 4041 & 105.7 & 1849 \\
\bottomrule
\end{tabular}
\caption{Latency and memory usage are measured during policy training and inference with batch size 64. Training latency is the average time per training step (forward + backward) over 100 iterations. Inference latency is the average time to process observations and sample an action chunk with 8 flow-matching steps, also averaged over 100 iterations.}
\label{table:policy_stats}
\end{table}

\begin{table}[t]
\centering
\begin{tabular}{lccc}
\toprule
\textbf{ViT} & \textbf{Tokens} & \textbf{Latency (ms)} & \textbf{GFLOPs} \\
\midrule
\textbf{Fine}     & 324 & 243.8 & 1905.4 \\
\textbf{Coarse}   & 20 & 17.6 & 126.9 \\
\textbf{Foveated} & 20 & \textbf{16.4} & \textbf{115.6} \\
\bottomrule
\end{tabular}
\caption{Comparison of ViT inference using different patch tokenization patterns in terms of token count, latency, and GFLOPs, evaluated with a batch size of 64. Latency is averaged over 100 iterations.}
\label{table:vit_stats}
\vspace{-10pt}
\end{table}

Table~\ref{table:policy_stats} reports latency and memory usage during training and inference. The largest gap appears in training: \textit{Fine} is nearly 8× slower and uses 5× more GPU memory than other methods, driven by its much higher ViT token count (324 vs. 20 patches). During inference, differences are smaller because the flow matching transformer runs multiple sampling steps (8 in our case) while image features are processed only once by the ViT. Nonetheless, \textit{Fine} still shows ~3× higher latency at batch size 64 compared to other methods. Memory usage during inference is similar across methods, as no backpropagation is performed. Between gaze methods, \textit{Fov-UNet} is slightly slower and more memory-intensive than \textit{Fov-Act} due to its additional UNet.

Table~\ref{table:vit_stats} further isolates the ViT encoder analyzing its latency and computation associated with \textit{Fine}, \textit{Coarse}, and \textit{Foveated} tokenization. At a batch size of 64, both latency and GFLOPs are significantly higher for the \textit{Fine} pattern, with the relative differences roughly corresponding to the token counts of each method. Although \textit{Foveated} and \textit{Coarse} use the same number of patches, \textit{Foveated} is faster because its smaller patch size reduces the patch embedding computation.

\subsection{Success Rates}

\begin{table*}[t]
\centering
\begin{tabular}{c|l | c c c c | c c c c}
\toprule
\multicolumn{1}{c}{} & & \multicolumn{4}{c|}{\textbf{No Pretraining}} & \multicolumn{4}{c}{\textbf{With MAE Pretraining}} \\
\cmidrule(lr){3-6} \cmidrule(lr){7-10}
 \multicolumn{1}{c}{} & \textbf{Task} & \textbf{Fine} & \textbf{Coarse} & \textbf{Fov-Act} & \textbf{Fov-UNet} & \textbf{Fine} & \textbf{Coarse} & \textbf{Fov-Act} & \textbf{Fov-UNet} \\
\midrule
\multirow{6}{*}{\rotatebox[origin=c]{90}{\textbf{Standard}}} 
& \texttt{CubeTransfer}   & 72 & 98 & 68 & \textbf{100} & \textbf{100} & \textbf{100} & 90 & \textbf{100} \\
& \texttt{PegInsertion}   & 26 & \textbf{32} & 26 & \textbf{32} & \textbf{40} & 32 & 38 & 32 \\
& \texttt{SlotInsertion}  & 56 & 57 & \textbf{60} & \textbf{60} & 57 & 66 & 64 & \textbf{70} \\
& \texttt{HookPackage}    & 28 & 18 & 12 & \textbf{56} & \textbf{68} & 56 & 30 & 57 \\
& \texttt{PourTestTube}   & 34 & 60 & 78 & \textbf{84} & 68 & 78 & 80 & \textbf{92} \\
& \texttt{ThreadNeedle}   & 57 & 48 & 62 & \textbf{74} & 84 & 66 & 74 & \textbf{92} \\
\midrule
\multirow{6}{*}{\rotatebox[origin=c]{90}{\textbf{Distractors}}} 
& \texttt{CubeTransfer}   & 46 & \textbf{76} & 68 & 66 & 66 & 36 & \textbf{68} & 60 \\
& \texttt{PegInsertion}   & 12 & 18 & 26 & \textbf{28} & 10 & 22 & \textbf{34} & 22 \\
& \texttt{SlotInsertion}  & 54 & 44 & \textbf{56} & 52 & \textbf{64} & 54 & 57 & \textbf{64} \\
& \texttt{HookPackage}    & 32 & 24 & 14 & \textbf{54} & 52 & 30 & 28 & \textbf{57} \\
& \texttt{PourTestTube}   & 12 & 30 & \textbf{68} & 38 & 32 & 34 & \textbf{46} & 30 \\
& \texttt{ThreadNeedle}   & 24 & 32 & \textbf{56} & \textbf{56} & 48 & 40 & 68 & \textbf{70} \\
\bottomrule
\end{tabular}
\caption{Success rates (\%) on simulation tasks comparing policies using different ViT patch tokenization schemes: \textit{Fine}, \textit{Coarse}, and \textit{Foveated} (Fov). For the \textit{Foveated} scheme, two gaze estimation methods are evaluated: \textit{Fov-Act} (end-to-end) and \textit{Fov-UNet} (two-stage). Policies are evaluated on tasks both with and without distractors (\textit{Standard} and \textit{Distractors}), and under two training settings: training the ViT from scratch (\textit{No Pretraining}) and fine-tuning a pretrained ViT (\textit{With MAE Pretraining}).}
\label{table:sim_results}
\vspace{-10pt}
\end{table*}

\begin{table}[t]
\centering
\begin{tabular}{l | c c | c c }
\toprule
\multicolumn{1}{c}{} & \multicolumn{2}{c}{\textbf{Standard}} & \multicolumn{2}{c}{\textbf{Distractors}} \\
\cmidrule(lr){2-3} \cmidrule(lr){4-5}
\textbf{Task} & \textbf{Fine} & \textbf{Fov-UNet} & \textbf{Fine} & \textbf{Fov-UNet} \\
\midrule
\texttt{Ball} & \textbf{64} & 62 & 48 & \textbf{56}  \\
\texttt{Toothbrush}  & \textbf{24} & 18 & \textbf{18} & 14 \\
\bottomrule
\end{tabular}
\caption{Success rates (\%) on real-robot tasks comparing \textit{Fine} policy (uniform tokenization) and \textit{Fov-UNet} (foveated tokenization). Policies are trained with MAE-pretrained ViTs and data augmentations, and evaluated both without (\textit{Standard}) and with (\textit{Distractors}) distractors.}
\label{table:real_results}
\vspace{-10pt}
\end{table}

For simulation tasks, policy performance is evaluated every 3,000 training steps, for a total of 10 evaluations. At each checkpoint, the policy is rolled out 50 times in the simulation both with and without distractors, and performance is measured by task success rate. The highest success rate across all checkpoints is reported as the final performance. For real-robot tasks, only the final checkpoint after 30,000 steps is evaluated with 50 rollouts. Simulation results are shown in Table~\ref{table:sim_results} and real-robot results in Table~\ref{table:real_results}.

Notably, although our initial aim was to reduce computational complexity through foveation, as shown in Table~\ref{table:sim_results} the method also improves performance in many cases. When comparing simulation results with no ViT pretraining (i.e., training from scratch) in the in-distribution (\textit{Standard}) setting, \textit{Fov-UNet} consistently matches and often outperforms other methods in success rate. While \textit{Fov-Act} underperforms \textit{Fov-UNet} overall, it still surpasses the \textit{Fine} and \textit{Coarse} baselines on tasks like \texttt{ThreadNeedle} and \texttt{PourTestTube}.

Under the \textit{Distractors} setting without pretraining, the advantage of foveated tokenization becomes more pronounced, suggesting that foveation improves robustness to visual distractors. Interestingly, \textit{Fov-Act} outperforms \textit{Fov-UNet} on some tasks here, as distractors can misalign \textit{Fov-UNet}'s predicted gaze, particularly in tasks like \texttt{PourTestTube}.

With MAE pretraining in the Standard setting, all methods improve. The \textit{Fine} policy achieves top performance on two tasks, while \textit{Fov-UNet} still leads on three, doing better particularly for \texttt{ThreadNeedle} and \texttt{PourTestTube}. With MAE pretraining and distractors, foveated tokenization is overall better than uniform tokenization across tasks, highlighting its robustness.

On the \textit{HookPackage}, we note that \textit{Fov-Act} performs worse than all other methods. Here, the robot must shift its gaze from a package to a small hook in the periphery, which becomes difficult to track due to foveation’s downscaling of peripheral regions. \textit{Fov-Act} struggles to fix its gaze onto the hook, revealing a limitation of end-to-end gaze prediction.

For real-robot tasks (\texttt{Ball} and \texttt{Toothbrush}), differences between \textit{Fov-UNet} and \textit{Fine} are less pronounced than in simulation. \textit{Fine} slightly outperforms on \texttt{Toothbrush}, while \textit{Fov-UNet} is comparable or better on \texttt{Ball}, particularly with distractors. 

Overall, foveation methods (\textit{Fov-UNet} and \textit{Fov-Act}) perform on par with \textit{Fine} and \textit{Coarse} baselines while being more robust to distractors. In simulation, \textit{Fov-UNet} surprisingly even showed higher success rates overall, suggesting gaze could provide a useful cue for task success. These results highlight the benefits of foveated tokenization, which preserves detail at the gaze while dramatically reducing token count, indicating that focusing on select visual features is sufficient for effective and efficient task performance.  

\section{Conclusion and Future Work}

In this work, we present an IL framework that leverages gaze for effective and efficient bimanual manipulation. By integrating foveation into ViTs, our IL framework emulates human visual processing. We demonstrate that foveated ViTs can match or surpass standard ViTs in policy performance while providing greater robustness to visual distractors, alongside substantial reductions in training and inference time and space overheads. We compare two gaze estimation approaches: a two-stage method that separates gaze and action estimation, and an end-to-end method that concurrently generates gaze and action. While the two-stage approach generally outperforms, the end-to-end method offers reduced complexity. Additionally, we introduce and open-source the GIAVA setup, including two manipulation arms, an AV arm, and a gaze measurement and synchronization pipeline. We also provide the first simulation benchmark and dataset for IL with active vision and human gaze tracking. We believe that developing more human-like vision systems, capable of both searching and fixating on the most relevant information, is essential for advancing robot learning. Our results highlight that incorporating gaze enables more robust and efficient robot learning and perception.


Looking ahead, we aim to validate foveated policy learning across a broader range of tasks and identify which classes of manipulation (e.g., high-precision, long-horizon, contact-rich) benefit most from foveation. We also plan to integrate popular pretrained ViTs (like DINOv2~\cite{oquab2023dinov2}) with foveated patch tokenization using recent adapter-based methods for multiscale patches~\cite{choudhury2025accelerating}, as our MAE-pretrained ViTs remained sensitive to lighting variation. Overall, foveation remains an untapped inductive bias: as we move toward human-level dexterity, the value of the cues offered by human visual evolution and the associated computational savings and performance improvements will only grow.

\bibliographystyle{IEEEtran}
\bibliography{IEEEabrv,references}

\end{document}